\documentclass[letterpaper, 10 pt, conference]{ieeeconf}  

\IEEEoverridecommandlockouts                              

\usepackage{amsmath} 
\usepackage{cite}
\usepackage{mathtools}
\usepackage[hidelinks]{hyperref}
\usepackage{siunitx}
\usepackage{listings}
\usepackage{xcolor}

\usepackage{booktabs}

\definecolor{codegreen}{rgb}{0,0.6,0}
\definecolor{codegray}{rgb}{0.5,0.5,0.5}
\definecolor{codepurple}{rgb}{0.58,0,0.82}
\definecolor{backcolour}{rgb}{0.95,0.95,0.92}

\lstdefinestyle{mystyle}{
    backgroundcolor=\color{backcolour},   
    commentstyle=\color{codegreen},
    keywordstyle=\color{magenta},
    numberstyle=\tiny\color{codegray},
    stringstyle=\color{codepurple},
    basicstyle=\ttfamily\footnotesize,
    breakatwhitespace=false,         
    breaklines=true,                 
    captionpos=b,                    
    keepspaces=true,                 
    numbers=left,                    
    numbersep=2pt,                  
    showspaces=false,                
    showstringspaces=false,
    showtabs=false,                  
    tabsize=2,
    xleftmargin=8pt
}

\hypersetup{}

\title{\LARGE \bf
PneuDrive: An Embedded Pressure Control System and Modeling Toolkit for Large-Scale Soft Robots
}

\author{Curtis C. Johnson, Daniel G. Cheney, Dallin L. Cordon, Marc D. Killpack
\thanks{This material was based upon work supported by the National Science Foundation under Grant No. 1935312 and  2024792.}
\thanks{All authors are with the Robotics and Dynamics Laboratory at Brigham Young University in Provo Utah, USA.}
}

\begin{document}

\maketitle
\thispagestyle{empty}
\pagestyle{empty}

\begin{abstract}

In this paper, we present a modular pressure control system called PneuDrive that can be used for large-scale, pneumatically-actuated soft robots. The design is particularly suited for situations which require distributed pressure control and high flow rates. Up to four embedded pressure control modules can be daisy-chained together as peripherals on a robust RS-485 bus, enabling closed-loop control of up to 16 valves with pressures ranging from 0-100 psig (0-689 kPa) over distances of more than 10 meters. The system is configured as a C++ ROS node by default. However, independent of ROS, we provide a Python interface with a scripting API for added flexibility. We demonstrate our implementation of PneuDrive through various trajectory tracking experiments for a three-joint, continuum soft robot with 12 different pressure inputs. Finally, we present a modeling toolkit with implementations of three dynamic actuation models, all suitable for real-time simulation and control. We demonstrate the use of this toolkit in customizing each model with real-world data and evaluating the performance of each model. The results serve as a reference guide for choosing between several actuation models in a principled manner. A video summarizing our results can be found here: \url{https://bit.ly/3QkrEqO}.
\end{abstract}

\section{INTRODUCTION}

Pneumatics are a common method of actuation for soft robots. Coupled with the inherent compliance of the soft robot structure, pneumatics can often improve safety while the robot makes unplanned contact with the environment.

Many pneumatically actuated soft robots are small (i.e. centimeter scale \cite{Stolzle_Santina_2022, addressablePCB}). As the size of soft robots increase, many can no longer support their own weight or are only capable of carrying small payloads. Due to their tremendous potential for flexibility and adaptability, we aim to create solutions that facilitate the `scaling up' of soft robots (i.e. meter scale). As pneumatically-actuated soft robots scale up in size and payload capability, their actuators will require higher flow rates, will exert higher forces at higher pressures, and will more directly affect the dynamic response of the system. Industrial pneumatic actuators and controllers that can handle such requirements are unsuitable for soft robots due to their size and weight. This suggests a need for a small, lightweight, and affordable pressure control system that can be distributed over large physical distances. 

\begin{figure}[ht]
    \centering
    \includegraphics[width=.5\textwidth]{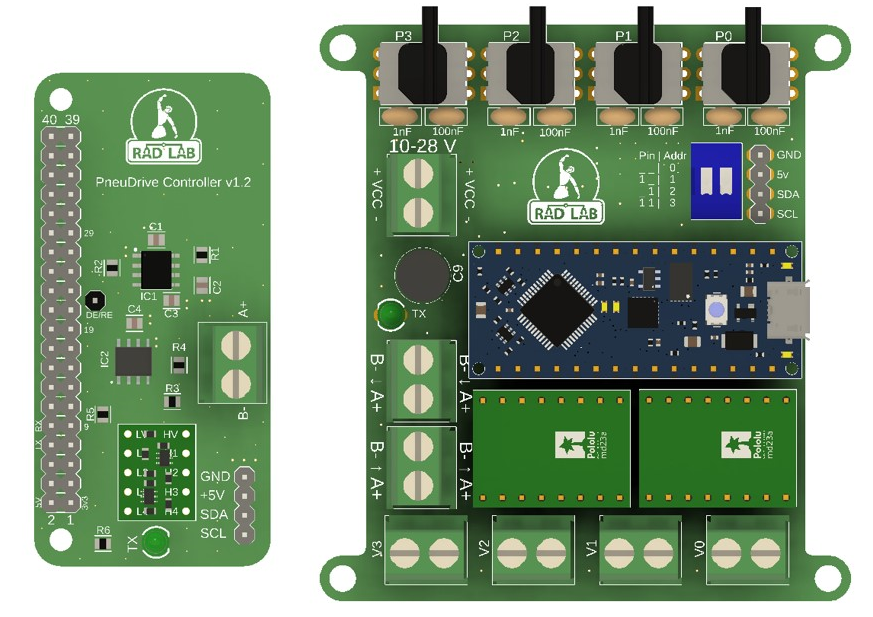}
    \caption{PneuDrive Controller (left) and PneuDrive Embedded (right) boards.}
    \label{fig:PneuDrive Boards}
\end{figure}

In addition, as more dynamic tasks are attempted with large-scale soft robots, simulating the effect and transient response of pneumatic actuators will play a more important role. Existing high-fidelity multi-physics simulation tools are too slow and computationally expensive to meet the demands of real-time model-based control (MBC) or reinforcement learning (RL). As a result, actuator dynamics have largely been ignored in modeling and control of pneumatically-driven soft robots \cite{Stolzle_Santina_2022}. However, recent efforts have begun to incorporate them in various ways, suggesting a need for a principled methodology to choose between various models.

As a step in addressing these needs, this paper makes two main contributions: 
\begin{itemize}
    \item We present an open-source pressure control system called PneuDrive, designed specifically for large-scale (meter-length) soft robots and demonstrate its capabilities in hardware. The total cost of the system is roughly \$200 USD per embedded board. Designs and software can be found here: \url{https://bit.ly/463MDUI}.
    \item We provide an easy-to-use software toolkit with implementations of three different dynamic actuation models originally derived in \cite{andersen1967analysis, springerpneumatic, best2016new, Tassa_Wu_Movellan_Todorov_2013}. These models are all suitable for real-time simulation and control, but vary in terms of performance characteristics. We use this toolkit to evaluate the trade-offs between each type of model, focusing specifically on speed, accuracy, and ease of system identification. The findings we present can serve as a reference guide to choosing a suitable actuation model for a particular application. This toolkit can be found here: \url{https://bit.ly/3s8UOkG}
\end{itemize}

Our requirements driving this work are as follows:
\begin{enumerate}
    \item Large Scale - We need the system to regulate pressures of up to 80 psi while providing high-flow rates with proportional flow valves.
    \item Distributed control - We need to control at least 12 valve/chamber pairs separated by physical distances of about one meter with minimal wiring and tubing.
    \item Expandable - Ideally the system is capable of incorporating additional sensors for future work (e.g. IMUs).
    \item Easy to use - The solution should use low-cost, off-the-shelf components that are easy to replace and the software should be open-source and easy to customize.
    \item Communication - The system must easily connect to other devices with low latency and reliable communication for real-time control loops.
    \item Simulation - We need models that capture the strong dynamic nonlinearites and coupling with pneumatics. Since we intend on using these models for MBC and RL they must be reasonably accurate and real-time capable.
\end{enumerate}

Next we explore current state-of-the-art pressure control solutions with respect to our design requirements and summarize the existing work towards modeling pneumatic actuator dynamics.

\subsection{Related Work}
The Soft Robotics Control Unit (SRC) \cite{DesktopSRC} is a solution based on a Raspberry Pi which can stack up to six pressure control hats for controlling up to 12 pneumatic channels. The authors provide APIs for MATLAB, Simulink, and Unity, with TCP/IP communication possible with minimal software modifications. The Pneumatic Box \cite{PneumaticBox} is a similar solution using a BeagleBone Black (BBB) that integrates well with common robotics middleware (e.g ROS). PneuSoRD \cite{PneuSoRD} uses microcontroller devices like the Arduino Due instead of single board computers (SBC) like the Raspberry Pi or BBB and can control up to five proportional valves. Each design fits our needs in all areas \emph{except} distributed control. The design of these platforms all require the valves and pressure sensors to be located centrally, multiplying the amount of tubing and wiring needed to operate all 12 valves. 

The authors of \cite{addressablePCB} released an innovative design that uses two miniature ON-OFF valves to mimic proportional valve control. Each board is individually addressable on an I2C bus allowing control of up to 127 pseudo-proportional valves. While ideal for small soft robots, the miniature valve size severely limits flow rates. Software APIs are not provided but could be added without much difficulty. However, the I2C protocol was not designed to be used over long physical distances and is sensitive to electromagnetic interference. The Wireless Compact Control Unit (WICCU) \cite{WiCCU} offers higher flow rates but limits control to three chambers due to six integrated valves. Its Bluetooth communication reduces wiring complexity, but is known to have latency on the order of 10 ms \cite{bluetoothLatency}, which may cause instability with high-bandwidth real-time control loops. FlowIO \cite{flowio} is small and compact pressure control solution with several communication APIs and many customization options. As it is designed for small scale soft robots and wearables, the pressure range (-30 to 30 psi) and flow rates (up to 3.2 L/min) are unfortunately not sufficient for our applications.

Related to modeling the dynamics of these pneumatic actuators, foundational texts like \cite{andersen1967analysis} and \cite{springerpneumatic} derive models for pneumatic chamber pressures of variable volume, but lack specifics on mass flow rate and valves. Existing models like ISO 6358 and manufacturer-provided valve models attempt to model flow regimes and orifices but require extensive system identification \cite{quarterellipseISO}. While nonlinear models are successful in offline soft robotics simulations \cite{Joshi_Paik_2021}, real-time hardware implementation demands more efficient methods. Approaches like parameterized models \cite{Tassa_Wu_Movellan_Todorov_2013} show promise, as do simplified first-order pressure dynamics \cite{best2016new}.

The remainder of this paper is organized as follows. Section \ref{sec:hardware} presents the physical design and specifications of PneuDrive and Section \ref{sec:modeling} presents our modeling toolkit and a brief overview of each dynamic model. Section \ref{sec:results} presents results from testing the PneuDrive in different configurations as well as an in-depth evaluation of each dynamic actuation model using our modeling toolkit.

\section{METHODS}

\subsection{Hardware Design}
\label{sec:hardware}

\subsubsection{Serial Communication}
\begin{figure}
    \centering
    \includegraphics[width=0.5\textwidth]{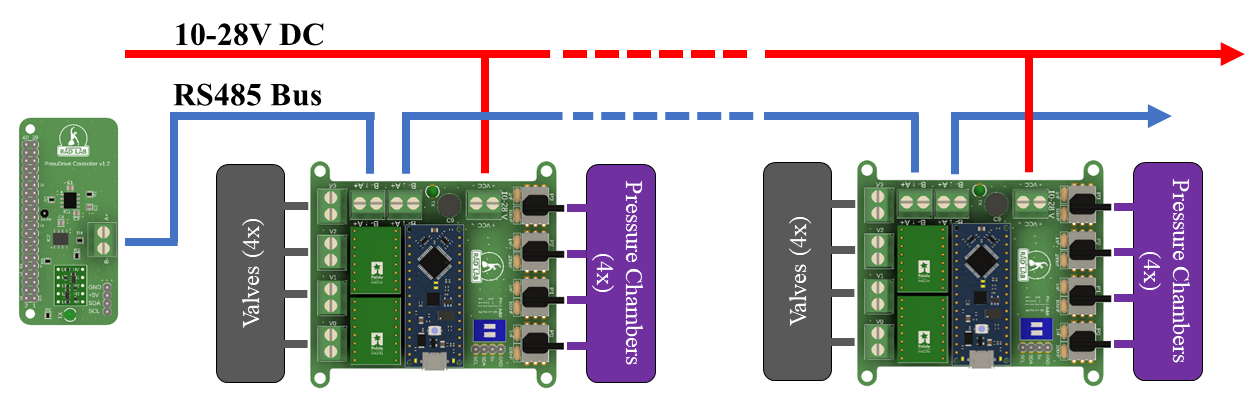}
    \caption{Two modules connected on a single RS485 bus for distributed embedded control of large-scale soft robots. Each board is powered from a common power bus (10-28V DC) for closed-loop control of up to 4 valve/chamber pairs. The dashed lines indicate that the bus can extend over large physical distances.}
    \label{fig:daisychain}
\end{figure}

PneuDrive operates in a serial bus configuration and is pin compatible with a number of single board computers (e.g. Raspberry Pi, ODROID, NVIDIA Jetson). The single board computer (SBC) should be capable of running ROS if the pressure control system will be used as a ROS node. Otherwise an SBC running Ubuntu is sufficient. For our application, we chose to use the ODROID N2+ running Ubuntu 20.04 and ROS Noetic. The controller is connected to a series of daisy-chained PneuDrive embedded boards with the two-wire RS485 bus (see Fig.~\ref{fig:daisychain}). Each embedded board is responsible for the closed-loop control of up to four valve/chamber pairs. The RS485 bus on PneuDrive offers tremendous flexibility to the user because of its robustness to electrical interference, its ability to function over large physical distances (theoretically a limit of 1200 meters), and its reliability features such as built-in electrostatic discharge (ESD), surge protection, and hot plug-in support.

\begin{figure}
    \centering
    \includegraphics[width=0.5\textwidth]{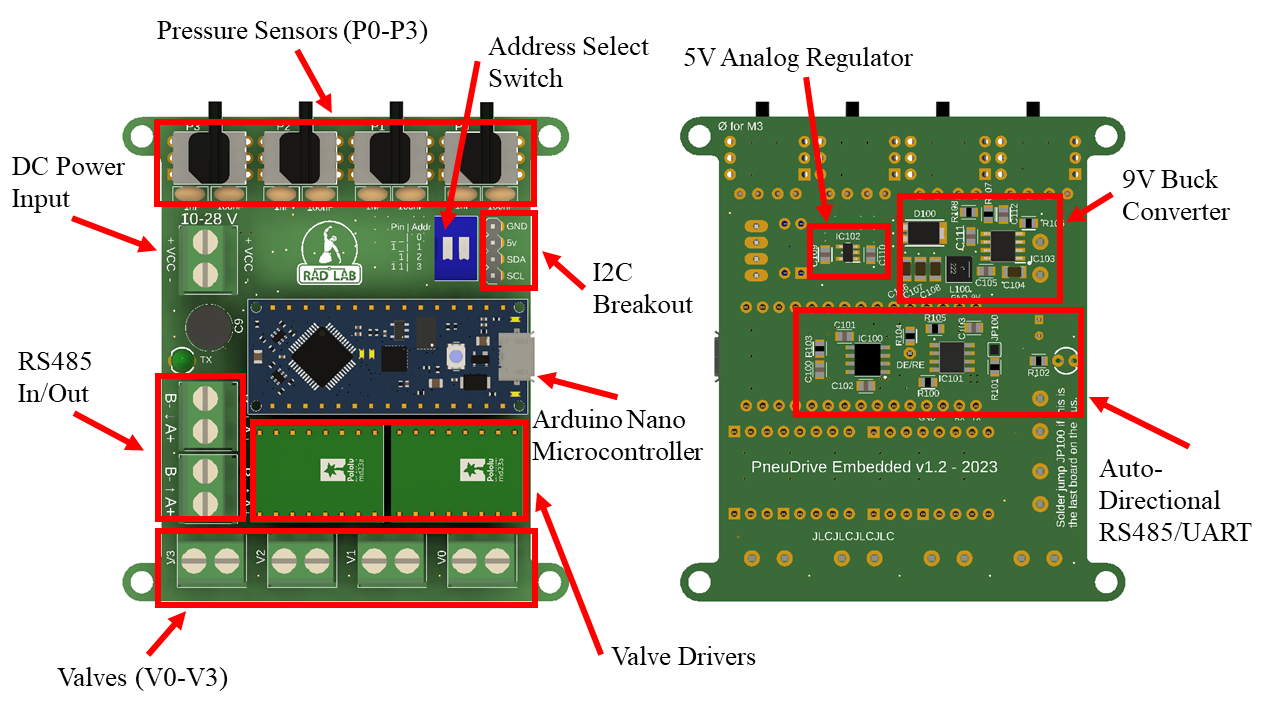}
    \caption{Onboard functionality of PneuDrive Embedded board.}
    \label{fig:single board diagram}
\end{figure}

\begin{figure}
    \centering
    \includegraphics[width=0.5\textwidth]{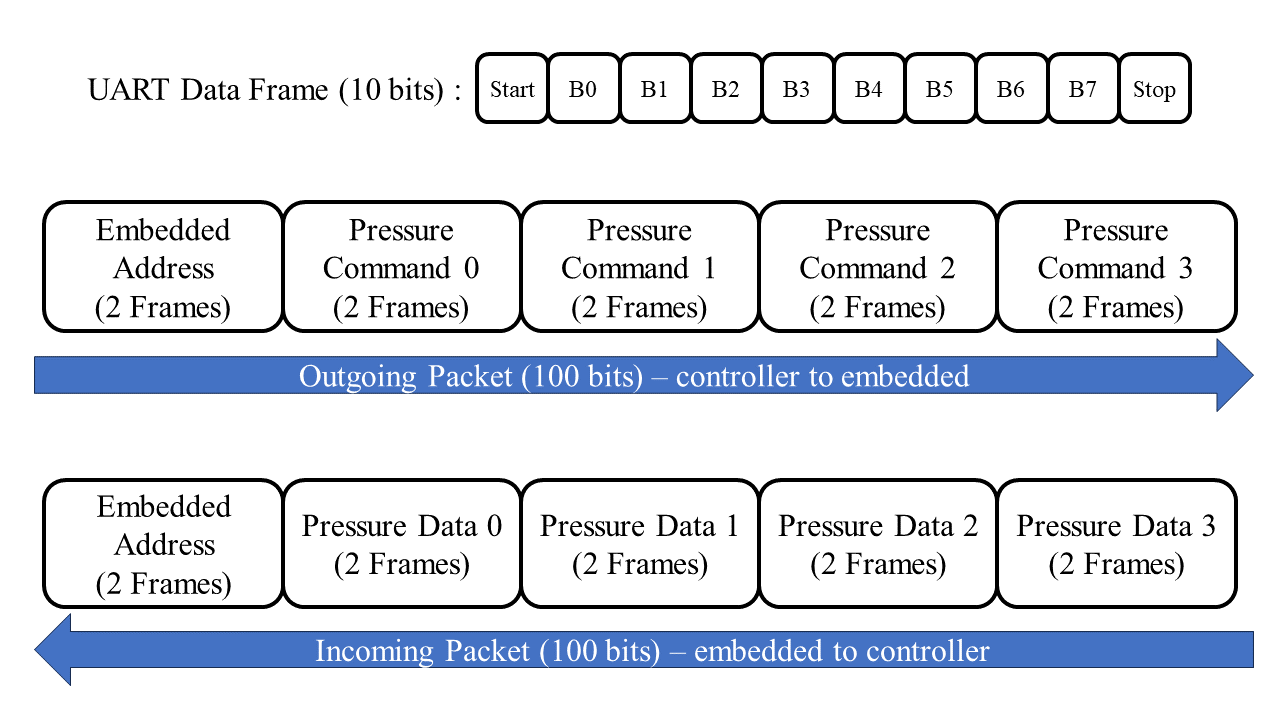}
    \caption{Communication Protocol for PneuDrive. Each data frame uses the `8N1' serial setting for a total of 10 bits per data frame. Each 16-bit integer requires two frames, for a total of 10 data frames per packet.}
    \label{fig:comm-protocol}
\end{figure}

The daisy-chained PneuDrive embedded boards are assigned unique addresses through the Address Select Switch (see Fig.~\ref{fig:single board diagram}). The switch allows up to four addresses on a single RS485 bus, though a theoretical maximum of 256 devices is possible if addresses are programmed in software. Device addresses start at 0xFFFF and decrement by one for each successive device, avoiding conflicts with 10-bit resolution ADC data which ranges from 0-1023 (0x0000 - 0x03FF). The communication protocol uses data packets composed of five 16-bit integers as shown in Fig.~\ref{fig:comm-protocol}. We use the common `8N1' serial configuration indicating eight data bits, no parity bit, and one stop bit (see UART Data Frame in Fig.~\ref{fig:comm-protocol}). The controller initiates a transmission with an outgoing packet containing pressure commands and the addressed embedded device responds with pressure data. The addresses are used as delimiters to avoid bus contention, ensuring  that only one device responds to outgoing packet and that it is disregarded by other devices on the bus.

The RS485 implementation in PneuDrive is half-duplex (meaning sending and receiving data cannot happen concurrently) and only requires two wires. Each PneuDrive embedded board defaults to `read' mode and only switches to `write' mode when responding to an outgoing packet sent from the PneuDrive controller (see Fig.~\ref{fig:comm-protocol}). This functionality is called `auto-direction' and its implementation in PneuDrive is based on a design released by Texas Instruments Inc. \cite{Instruments_2016a}. A 555 timer in monostable mode is used to generate a `write-enable' pulse for a given duration when the first bit of the incoming data packet is detected (i.e. the start bit of the first data frame). The duration is slightly longer than the transmission time of a full data packet. For example, at our chosen baud rate of 1 Mbps, each bit has a pulse width of 1 \unit{\micro \second}. Accordingly, one data packet takes 100 \unit{\micro \second} to transmit so we need to choose an appropriate `write-enable' pulse time, which is calculated as $t_{enable} = 1.1 R C$ (see 555 timer datasheet). The tightest tolerance for commercially-available capacitors is 5\%, so we use $C = .1 \unit{\micro \farad} \pm 5\%$. The resistance $R$ we require is the lowest commercially-available resistor that keeps $t_{enable} > 100 \unit{\micro \second}$ in the worst tolerance case. Using the propagation of error equation, $R = 976 \unit{\ohm} \pm 1\%$ meets our needs and ensures a $t_{enable} = 107 \pm 5 \unit{\micro \second}$.  The duration of the `write-enable' pulse can be easily modified by changing a single resistor. This feature simplifies the software required to avoid bus collisions and data corruption that can occur if devices enter `read' mode prematurely.

\subsubsection{Electrical Description}
Each PneuDrive Embedded board can be powered from a range of 10-28V DC to accommodate most pneumatic valves. The valves connected to the board (V0-V3) will be driven at the voltage of the DC Power Input (see Fig.~\ref{fig:single board diagram}). Both dual valve drivers are off-the-shelf breakout boards that actively limit current to the valves to 900 mA peak or 700 mA continuous. The current limit can be adjusted to anywhere between 100-1400 mA for a given application by changing a single current sense resistor on the driver board, although additional heat sinking is required for loads of more than 700 mA continuous. The drivers are plugged into a socket above the PCB so there is space for an additional heat sink if necessary.

An on-board buck converter regulates the DC power input to nine volts to power the microcontroller and its peripherals. The regulator can source up to 3.6 amps but the load from PneuDrive will typically be much smaller than that -- up to 1 amp. The microcontroller distributes its 5V logic throughout the board to the other chips, with the exception of a dedicated 5V low-noise low dropout regulator (LDO) for exclusive use of the analog pressure sensors (P0-P3).

Additionally, the required pins for a 5V I2C bus are broken out for an easy option to add additional I2C sensors to PneuDrive (e.g. temperature, IMU). The pins are compatible with most I2C connection systems like Sparkfun Qwiic \cite{Sparkfun} and Adafruit STEMMA QT \cite{STEMMA}.

The PneuDrive controller board does not have a dedicated power input because it does not need to power valves. It uses the 5V and 3.3V power pins available on the standard SBC pinout. The controller also carries a 3.3V/5V bidirectional logic level converter since most SBCs operate at a 3.3V logic level and the PneuDrive embedded boards operate at a 5V logic level. 

\subsubsection{Software Interface}

The source code to run PneuDrive is implemented in C++ for high performance, but we expose the system in software via two different interfaces via a ROS node or python scripts.

The PneuDrive ROS Node exposes each embedded device on the bus via rostopics. Each embedded device consists of a subscriber to a pressure command topic for incoming command packets and a publisher for outgoing data packets (see Fig.~\ref{fig:comm-protocol}). The user of the PneuDrive ROS node simply needs to publish commands to the pressure command topics and subscribe to the pressure data topics. 

The Python Scripting interface allows a user to access data in a Python script, without using ROS. We use pybind11 \cite{pybind11} to wrap the C++ source code into an importable module called \verb"pneudrive_py". The following is an example of how to use the API: 

\begin{lstlisting}[language=Python]
import numpy as np
from pneudrive_py import PressureController

uart_port = '/dev/ttyS1'
num_devices = 4
my_controller = PressureController(uart_port, num_devices)

#check communication with all expected devices
my_controller.ping_devices()

pressure_cmd = np.array([1,2,3,4])
for i in range(num_devices):
    my_controller.set_pressure_commands(i, pressure_cmd)
    data = my_controller.get_pressure_data(i)
\end{lstlisting}

\subsection{Pressure Dynamic Models}
\label{sec:modeling}
\begin{figure}
    \centering
    \includegraphics[width=0.45\textwidth]{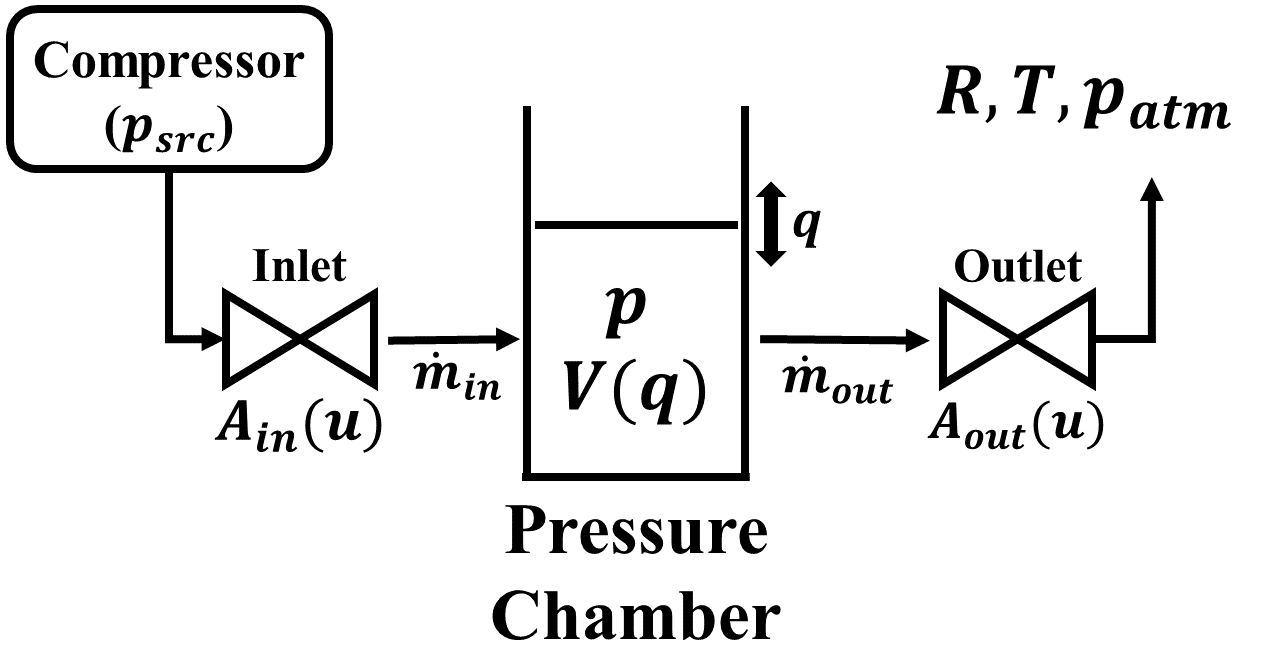}
    \caption{Functional diagram demonstrating the physical meaning of important variables in the pressure dynamic models for a pressure chamber of variable volume.}
    \label{fig:pressure_diagram}
\end{figure}

In this section, we provide a summary of each of the dynamic models implemented in our modeling toolkit. We chose these models in an effort to approximately span the types of models available in the literature in order to provide a useful reference guide for choosing an appropriate model, with a specific focus on system identification. Fig.~\ref{fig:pressure_diagram} is a functional diagram depicting physical significance of the variables used in this section.

\subsubsection{Linear Model}
The first model, which we refer to as the linear model, is simply a first order-linear ODE on the pressure $p$:
\begin{equation}
    \dot{p} = -\alpha p + \beta p_{cmd}.
\end{equation}

The authors of \cite{best2016new} chose this model based on observations from a step command to a PID pressure controller. They tuned $\alpha$ and $\beta$ manually with a Model Predictive Controller to improve performance. The linear model captures the fact that the input dynamics (i.e. valves/pressures) are not fast enough to be negligible. Such constraints are particularly important for model-based control and especially common for pneumatically-actuated soft robots. However, this model completely neglects the coupling between the joint-space dynamics and the input dynamics, which becomes substantial as soft robots become larger, more dynamic, or exert large forces on the environment. 

\subsubsection{Nonlinear Model}
The second model, which we refer to as the nonlinear model, is a first order, nonlinear ODE derived from first-principles. We refer the reader to several derivations of this model in \cite{andersen1967analysis, springerpneumatic} and summarize the results here.

The governing equation is 
\begin{equation}
\label{eq:nonlinear}
    \dot{p} = \gamma \frac{RT}{V(q)} (\dot{m}_{in} - \dot{m}_{out}) - \gamma \frac{w\dot{V}(\dot{q})}{V(q)}p
\end{equation}

where $\gamma$, $R$, and $T$ are constants, $V$ is volume, $w$ is a parameter to quantify volume model uncertainty, and $q$ is the joint angles of the robot.

For the mass flow rate, we choose
\begin{equation}
\label{eq:massflow}
    \dot{m} = A C_d \Psi(p_u, p_d) p_u \sqrt{\frac{2}{R T}}
\end{equation}

where $C_d$ is a unit-less discharge coefficient and accounts for losses stemming from a non-ideal orifice.

Note that by convention, mass $m$ flows from a higher-pressure upstream ($p_u$) to a lower-pressure downstream ($p_d$), resulting in
\begin{align}
    \dot{m}_{in} &= A_{in}(u) C_d \Psi(p_u=p_{src}, p_d=p) p_{src} \sqrt{\frac{2}{R T}} \\
    \dot{m}_{out} &= A_{out}(u) C_d  \Psi(p_u=p, p_d=p_{atm}) p \sqrt{\frac{2}{R T}}.
\end{align}

We approximate the function that maps control input $u$ to the effective orifice input/output area $A_{in/out}$ with the smooth approximation to the typical `V' shaped model for proportional valves presented in \cite{Tassa_Wu_Movellan_Todorov_2013}:

\begin{align}
    A_{in}(u)&=L_{in}+\operatorname{smax}\Bigl(B\left(u-U_{in}\right)-L_{in}\Bigr) \\
    A_{out}(u)&=L_{out}+\operatorname{smax}\Bigl(B\left(U_{out}-u\right)-L_{out}\Bigr)
\end{align}

\noindent where 

\begin{equation}
    \operatorname{smax}(x)  = \frac{\sqrt{x^2+1} + x}{2}.
\end{equation}
$u$ is the input to the system and $B$ scales the input units to area units. In the case of PneuDrive, the input takes the form of a voltage applied as a PWM signal to the motor drivers so $B$ has units of \unit{\square\meter\per\volt}. The input/output leakage of the valve in \unit{\square \meter} is $L$. $U$ corresponds to the input causing zero output (i.e. the input center) and has units of \unit{\volt}. To demonstrate closed-loop pressure command tracking, we use a simple proportional controller of the form

\begin{equation}
\label{eq:PID controller}
    u = k_p (p_{des} - p)
\end{equation}
though many other controllers could be implemented \cite{Stolzle_Santina_2022}.

The flow function $\Psi$ in (\ref{eq:massflow}) can take many forms. However, a common choice is ISO 6358:

\begin{equation}
\Psi (p_u, p_d) = \begin{cases}\psi_{\max } & \frac{p_{d}}{p_{u}} \leq b\text { (choked) } \\ \psi_{\max }\left[1-\left(\frac{\frac{p_{d}}{p_{u}}-b}{a-b}\right)\right]^{\beta} & \frac{p_{d}}{p_{u}}>b \text { (subsonic) }\end{cases}
\end{equation}

\noindent where 

\begin{equation}
    \psi_{\max }=\left(\frac{2}{\gamma+1}\right)^{\frac{1}{\gamma-1}} \sqrt{\frac{\gamma}{\gamma+1}}
\end{equation}
is just a constant.

The volume terms in (\ref{eq:nonlinear}) ($V$ and $\dot{V}$) are functions of the configuration variables $q$ and $\dot{q}$. The specific form of this model will vary depending on the application and is often uncertain in soft robots, hence the inclusion of the parameter $w$ in (\ref{eq:nonlinear}). We use the lengths derived in \cite{Allen_Rupert_Duggan_Hein_Albert_2020} for the cylindrical continuum joints we intend to use: 

\begin{equation}
\begin{aligned}
& l_0= h+r u \\
& l_1= h-r u \\
& l_2= h+r v \\
& l_3= h-r v. \\
\end{aligned}
\end{equation}

Note that $q = [u,v]$ and that the volume of the ith chamber is approximately $V_i = \pi \delta^2 l_i(q)$, where $\delta$ is the radius of the cylindrical pressure chamber.

$\dot{V}(q)$ is simply the time derivative of $V$

\begin{equation}
    \dot{V}_i(q) = \pi \delta^2 \frac{\partial l_i}{\partial q}\frac{\partial q}{\partial t}
    \label{eq:final-nonlinear}
\end{equation}

Together, (\ref{eq:nonlinear}-\ref{eq:final-nonlinear}) constitute the complete nonlinear model, with six parameters to fit: $L_{in}$, $L_{out}$, $B$, $U_{in}$, $U_{out}$, and $w$.

\subsubsection{Parametric Model}
The third model, which we call the parametric model, was originally presented in \cite{Tassa_Wu_Movellan_Todorov_2013}. While it is inspired by first-principles, there are no physical laws explicitly represented in this model: 

\begin{equation}
\begin{aligned}
\hat{u} & =u-c_1 \\
\bar{g} & =g\left(c_2 \hat{u}+c_3 \hat{u}^3\right) \\
s & =c_b+c_s \bar{g}+c_4 \dot{V} \\
\bar{k} & =k\left(\hat{u}, c_\gamma, c_9, c_8\right) \\
r & =\frac{c_7 \bar{k}+c_5 \dot{V}}{1+c_6 V} \\
\dot{p}(p, u, V, \dot{V} ; \mathbf{c}) & =(s-p) \times r
\end{aligned}
\end{equation}

The pressures $p$ in the parametric model are driven to a steady-state value $s$ and a rate value $r$. The parameters $c_1$-$c_9$ are the free parameters. Some of the parameters have physical meaning, but most exist to provide a degree-of-freedom to capture the correct function shape.

\subsubsection{Modeling Toolkit}
The modeling toolkit presented as part of this paper contains a Python module for each of these three model types. Each module contains an implementation of the model and a jupyter notebook with an example call to LMFIT \cite{lmfit} for the user to easily identify the parameters for each model and compare the results. The API allows the user to save identified parameters and use them for open-loop prediction or simulation of the system.

\section{RESULTS AND DISCUSSION}
\label{sec:results}

\subsubsection{PneuDrive Hardware Experiments}

\begin{figure}
    \centering
    \includegraphics[width=0.45\textwidth]{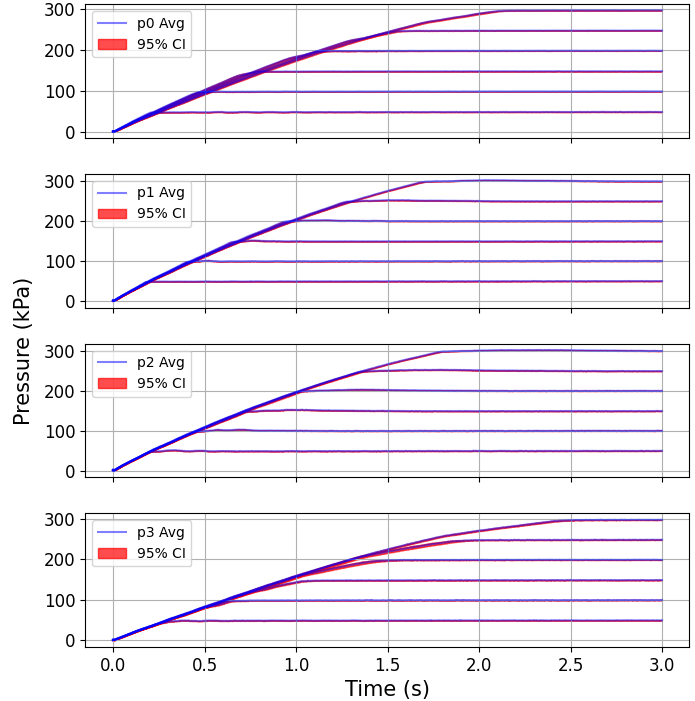}
    \caption{Pressure response characteristics of a four-chamber pneumatic joint to step commands from 50 to 300 kPa. Each response reflects the mean result of ten trials with a 95\% confidence interval over those same 10 trials.}
    \label{fig:stepResponseTest}
\end{figure}

\begin{figure*}
    \centering
    \includegraphics[width=\textwidth]{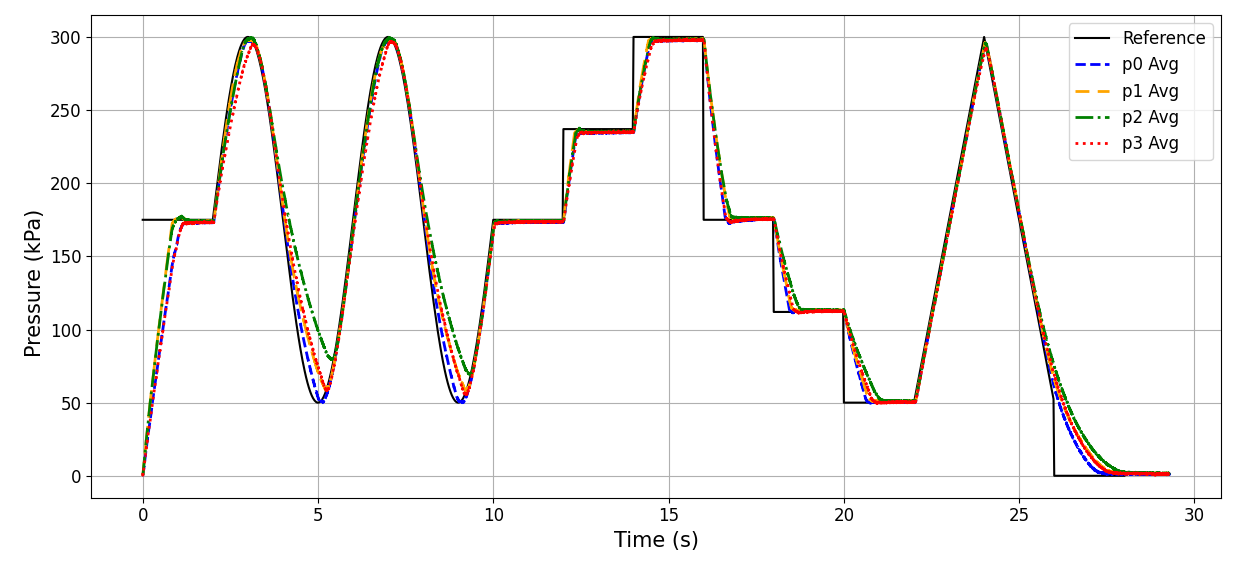}
    \caption{Pressure trajectory tracking with a four-chamber pneumatic joint. Commands range from 50 kPa to 300 kPa. Each response reflects the mean result of ten trials. The 95\% confidence interval over those ten trials, due to its minimal visibility, has been omitted for clarity.}
    \label{fig:trajectoryResponse}
\end{figure*}

To test PneuDrive, we evaluated step response characteristics and trajectory tracking for a single-joint of four pressure chambers (see \url{https://bit.ly/3QkrEqO}). The closed-loop step response and tracking performance are shown in Fig.~\ref{fig:stepResponseTest} and Fig.~\ref{fig:trajectoryResponse} respectively. While the performance is excellent for our application, there are clear differences between the responses for each chamber. This is indicative of normal variations associated with the joint hardware and volume-coupled dynamics. Our proportional controller (\ref{eq:PID controller}) does not attempt to counteract this, but it is worth noting that the models presented in this paper offer solutions for capturing these effects that are present on the physical hardware.

\begin{figure}
    \centering
    \includegraphics[width=0.7\columnwidth]{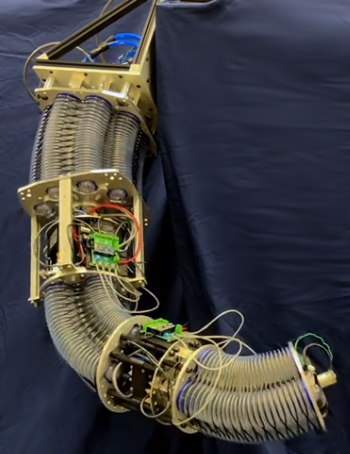}
    \caption{PnueDrive implemented on a large-scale pneumatic robot 1.16m in length. The arm consists of three embedded devices, each controlling four pressure chambers.}
    \label{fig:pneudrive_wholearm}
\end{figure}

PneuDrive embedded devices can easily be daisy-chained together. To demonstrate this, we implemented PneuDrive on a 1.16-meter-long pneumatic robot arm with three joints and 12 pressure-controlled chambers, shown in Fig.~\ref{fig:pneudrive_wholearm}. The video linked above shows the arm following a smooth, arbitrarily-chosen trajectory of commanded pressures. Table~\ref{tab:rates} demonstrates the average control rates we achieved with more embedded devices on the bus.

\begin{table}[]
    \centering
    \caption{Mean loop rate and standard deviation over 2044 iterations.}
    \label{tab:rates}
    \begin{tabular}{@{}ccc@{}}
        \toprule
        Number of Devices & Mean (Hz) & Standard Deviation (Hz) \\ \midrule
        1 & 1164.3 & 37.5 \\
        2 & 980.9  & 19.9 \\
        3 & 749.5  & 53.0 \\ \bottomrule
    \end{tabular}
\end{table}

\subsubsection{Modeling Experiments}
\begin{figure}
    \centering
    \includegraphics[width=\columnwidth]{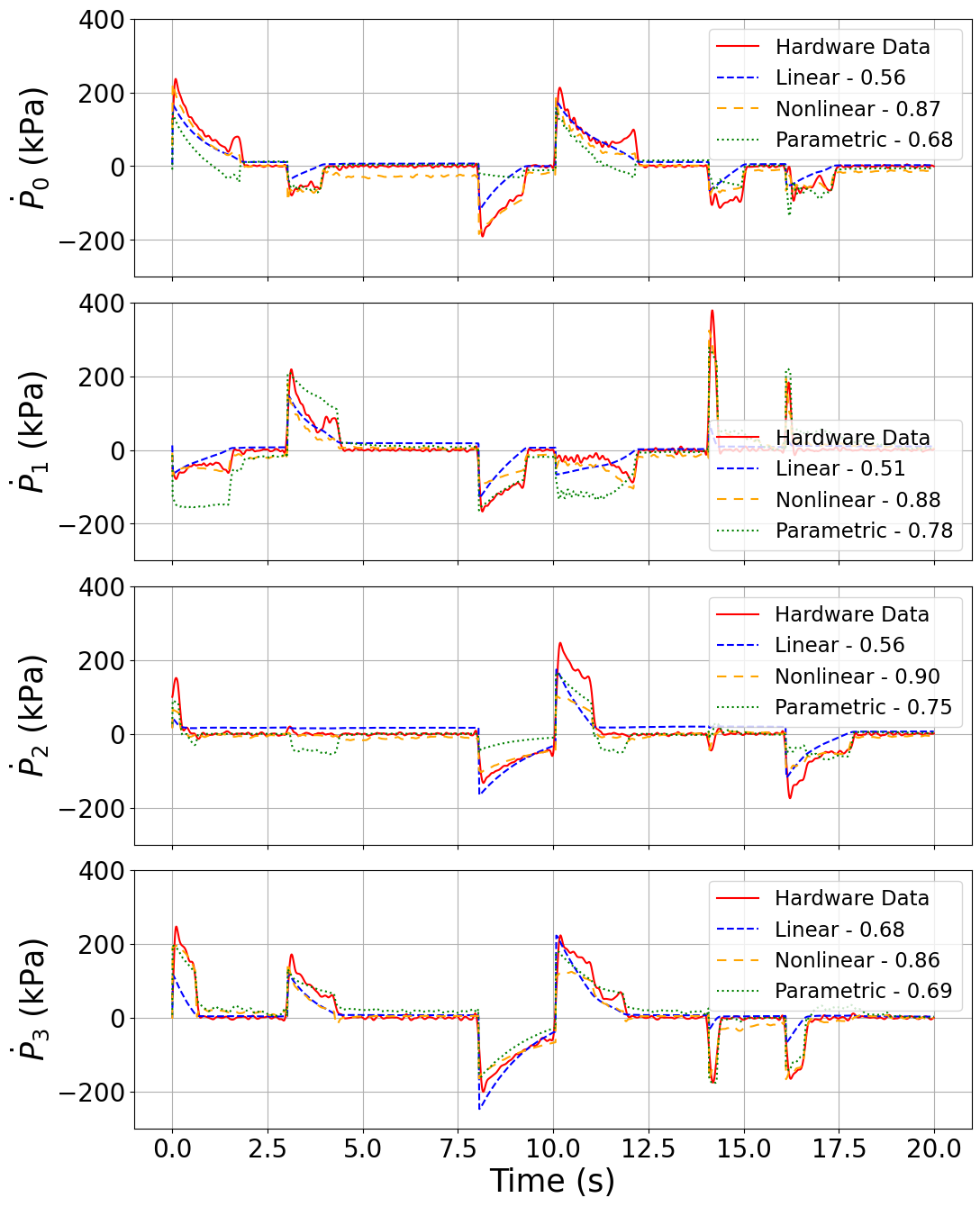}
    \caption{Comparison of models for each chamber in open-loop prediction of $\dot{p}$ for 20 seconds of validation data}
    \label{fig:modelComparison}
\end{figure}

In this section, we present the results of using our modeling toolkit to find optimal parameters on a single joint with four chambers. We optimize the parameters using 20,000 data points (200 seconds) of real-time data as a training data set and 2,000 data points (20 seconds) as our validation data set. The validation trajectories (see Fig.~\ref{fig:modelComparison}) are open-loop predictions of $\dot{p}$ given only the input signals for each model. To obtain the parameters we initialize 20 optimizations with random initial conditions for each model and keep the best set of parameters indicated in the legend by the highest $R^2$ value. These results are shown in  Fig.~\ref{fig:modelComparison}. Table~\ref{tab:iae_openloop_modelcomparison} has the Integrated Absolute Error (IAE) for each model over the 20 seconds of validation data. The evaluation time is averaged across the chamber models (C0 - C3). For each model type, a single optimization run converges in under one second, enabling fast generation of unique parameter sets for each chamber/valve pair, as shown in Fig.~\ref{fig:modelComparison}.

In summary, we found that the nonlinear model typically outperforms the linear and parametric models in accuracy. It consistently converges to a good set of parameters and captures transient dynamics while maintaining low steady state error at the cost of increased data processing and modeling work. The parametric model also captures transient dynamics without requiring extra data processing and modeling. The parametric model optimization did not converge consistently and typically exhibited worse steady state error. About 1/10 of the parametric optimizations converged to a usable model. The linear model is by far the simplest to process data for. It captures general trends consistently, as evidenced by the second-lowest IAE in Table \ref{tab:iae_openloop_modelcomparison}, and has the lowest execution time, all at the cost of model accuracy.

\begin{table}[]
    \centering
    \caption{IAE values for open-loop prediction for each chamber (C0 - C3) model found in Fig.~\ref{fig:modelComparison} in GPa.}
    \label{tab:iae_openloop_modelcomparison}
    \begin{tabular}{@{}cccccc@{}}
        \toprule
        Model & C0 & C1 & C2 & C3 & Eval Time (ms)\\ \midrule
        Linear & 41.13 & 44.80 & 45.16 & 39.77 & 0.111 \\
        Non-Linear & 35.84 & 37.43 & 27.51 & 31.56 & 0.628 \\
        Parametric & 48.67 & 67.81 & 46.29 & 66.17 & 0.432 \\ \bottomrule
    \end{tabular}
\end{table}

\section{CONCLUSION}
\label{sec:conclusion}
We have presented PneuDrive as a modular, open-source pressure control system tailored for large-scale pneumatically-actuated soft robots. We have also demonstrated the system on a distributed, large-scale soft robot. We presented a modeling toolkit to facilitate system identification and model comparison between several actuation models useful for simulation. Together, these attributes enable future research for distributed, large-scale soft robot hardware and simulation applications.






\section*{ACKNOWLEDGMENT}

The authors thank Phillip Hyatt, Alexa Lindberg, Brianna Taylor for their contributions during early prototypes of this project.



\bibliographystyle{IEEEtran.bst}
\bibliography{main}

\end{document}